# A K-means-based Multi-subpopulation Particle Swarm Optimization for Neural Network Ensemble

Hui Yu
*School of Computer Science and Technology*
*Zhejiang University of Technology*
Hangzhou, China
E-mail: 2111712025@zjut.edu.cn

Weiguo Sheng
*Department of Computer Science*
*Hangzhou Normal University*
Hangzhou, China
E-mail: w.sheng@ieee.org

*Abstract*—This paper presents a *k*-means-based multi-subpopulation particle swarm optimization, denoted as KMPSO, for training the neural network ensemble. In the proposed KMPSO, particles are dynamically partitioned into clusters via the *k*-means clustering algorithm at every iteration, and each of the resulting clusters is responsible for training a component neural network. The performance of the KMPSO has been evaluated on several benchmark problems. Our results show that the proposed method can effectively control the trade-off between the diversity and accuracy in the ensemble, thus achieving competitive results in comparison with related algorithms.

## I. INTRODUCTION

Neural network (NN) ensemble is a learning paradigm of creating a set of NNs, which are then trained and combined, as opposed to create just one single NN. It originates form Hansen and Salamon's work [1], which indicates that the generalization performance of a neural system can be significantly improved through ensembling a number of NNs. Many methods, such as [2] and [3], have shown that NN ensemble behaves remarkably well. This is due to the complementary among NN individuals for the same task.

Both theoretical and empirical studies [4-7] have indicated that the performance of NN ensemble depends greatly on the accuracy as well as the diversity of NN individuals. Liu and Yao [8] proposed a negative correlated learning algorithm to simultaneously train NN individuals in an ensemble. This is achieved by introducing a correlation penalty term to the error function of each individual such that individuals can interactively and cooperatively learn different aspects of training examples, thus producing negatively correlated errors among NN individuals. Zhou et al. [9] showed that it may be better to ensemble some component NNs rather than all the available NNs, and proposed a GA-based selection method (GASEN) to effectively select a subset from all trained individual NNs. Chandra et al. [10] developed a diverse and accurate ensemble learning algorithm (DIVACE) which is based on a multi-objective evolutionary algorithm. DIVACE tries to control the trade-off between diversity and accuracy in the ensemble by simultaneously optimizing two objective functions (i.e., empirical error function and correlation penalty function).

In this paper, we treat the ensembling NNs as a multimodal optimization problem, and propose a *k*-means-based multi-subpopulation particle swarm optimization, denoted as KMPSO, for training the NN ensemble. In the proposed method, particles are dynamically partitioned into clusters using the *k*-means clustering algorithm at every iteration, and each cluster is responsible for training a component NN. By employing dynamic cluster training, the proposed method is able to effectively control the trade-off between the diversity and accuracy in the ensemble. The performance of the proposed method has been evaluated on several benchmark problems and compared with related methods. The results show that the proposed method can deliver satisfactory NN ensembles and outperform related methods.

## II. NEURAL NETWORK ENSEMBLE

A widely accepted definition of NN ensemble is given by Sollich *et al.* [11]. Suppose the task is to use an ensemble comprising $N$ component NNs to approximate a function $F$: $R^m \rightarrow R^n$, and the predictions of component NNs are combined through weighted averaging where a weight $\omega_i$ ($i=1,2,\ldots,N$) satisfying (1) is assigned to the $i^{th}$ component NN

$$\sum_{i=1}^{N} \omega_i = 1, \ 0 \leq \omega_i \leq 1. \qquad (1)$$

The $k^{th}$ output variable of the ensemble is determined according to (3) where $f_{i,k}$ is the $k^{th}$ output variable of the $i^{th}$ component NN

$$F_k = \sum_{i=1}^{N} \omega_i f_{i,k}. \qquad (2)$$

Suppose $x \in R^m$ is sampled according to a distribution $p(x)$, the expected output of $x$ is $t(x)$, and the actual output of the $i^{th}$ component NN is $f_i(x)$. Then the output of the ensemble on $x$ is:

```
procedure KMPSO
    Initialize particles with random positions and velocities.
    Set particles' pbests to their current positions.
    Calculate particles' fitness.
    for step t=0→T-1 do
        Cluster particles' pbests using the k-means algorithm.
        Update particles' velocities according to (16).
        Update particles' positions
        Recalculate particles' fitness.
        Update particles' pbests.
    end for
    Cluster particles' pbests using the k-means algorithm.
end procedure
```

Fig. 1. The procedure of KMPSO

$$F(x) = \sum_{i=1}^{N} \omega_i f_i(x). \quad (3)$$

The generalization error of the $i^{th}$ component NN and its associated ensemble, i.e., $E_i$ and $E$, on the distribution $p(x)$ are:

$$E_i = \int p(x)(f_i(x) - t(x))^2 dx, \quad (4)$$

$$E = \int p(x)(F(x) - t(x))^2 dx. \quad (5)$$

The weighted average of generalization errors of all component NNs can be expressed as:

$$\overline{E} = \sum_{i=1}^{N} \omega_i E_i. \quad (6)$$

The difference on the $i^{th}$ component NN is defined as:

$$D_i = \int p(x)(f_i(x) - F(x))^2 dx. \quad (7)$$

The weighted average of the differences of all component NNs of the ensemble can be expressed as:

$$\overline{D} = \sum_{i=1}^{N} \omega_i D_i. \quad (8)$$

Then, as pointed out by Krogh *et al.* [12], the generalization error of ensemble can be written as:

$$E = \overline{E} - \overline{D}. \quad (9)$$

From the above equation, we can see that the increase of the differences will induce the decrease of generalization error of the ensemble, supposing the generalization error of component NNs is not increased. This inspires us to design a dynamic clustering training method to preserve the diversity of component NNs.

## III. PARTICLE SWARM OPTIMIZATION

Particle swarm optimization (PSO) [13] algorithm is an adaptive algorithm based on the simulation of birds flocking or fish schooling looking for food. The swarm in a $D$-dimensional space, where each particle represents a candidate solution to the optimization problem. Particles' flight is influenced by the best positions previously found by themselves and other particles. Based on this, the position and velocity for the $i^{th}$ particle is represented as $\vec{x}_i = (x_{i1},...x_{id},...x_{iD})$ and $\vec{v}_i = (v_{i1},...v_{id},...v_{iD})$ respectively, its individual best position is expressed as $\vec{p}_i = (p_{i1},...p_{id},...p_{iD})$, which is called *pbest*. $\vec{p}_g$ records and represents the best position of all particles experienced in the swarm, which is called *gbest*. Generally, the updating equations of velocity and position of the particle are given as follows:

TABLE I. FIVE BENCHMARK CLASSIFICATION PROBLEMS.

| Data set | No. of examples | No. of attributes | No. of classes |
|---|---|---|---|
| Australian credit card | 690 | 14 | 2 |
| Diabetes | 768 | 8 | 2 |
| Glass | 214 | 9 | 6 |
| Breast cancer | 699 | 9 | 2 |
| heart | 303 | 13 | 2 |

$$v_{id} = wv_{id} + c_1 r_1 (p_{id} - x_{id}) + c_2 r_2 (p_{gd} - x_{id}), \quad (10)$$

$$x_{id} = x_{id} + v_{id}, \quad (11)$$

where $w$ is an inertia weight, $c_1$ and $c_2$ are acceleration constants, $r_1$ and $r_2$ are random values generated following the uniform distribution between 0 and 1.

## IV. PROPOSED METHOD

In this section, a *k*-means-based multi-subpopulation particle swarm optimization is proposed for training the NN ensemble. We use a dynamic cluster training method to control the trade-off between the diversity and accuracy in the ensemble. The details of the component NN model, encoding scheme, fitness function and the procedure of proposed method are as follows.

### A. The Component Neural Network

In the proposed method, each component NN is a typical three-layer feedforward NN, where $X (x_1, x_2, ..., x_n)$ and $F(f_1, f_2, ..., f_m)$ are inputs with $n$ elements and outputs with $m$ nodes, respectively. The output values of nodes in the hidden layer and in the output layer can be defined as:

$$h_j = S\left(\sum_{i=1}^{n} w_{ji} x_i + b_j\right), \quad 1 \le j \le q \quad (12)$$

and

$$f_k = S\left(\sum_{j=1}^{q} w_{kj} h_j + b_k\right), \quad 1 \le k \le m, \quad (13)$$

respectively. Here, $S$ is a sigmoid function, $w_{ji}$ denotes the connection weights between input nodes and hidden nodes, $w_{kj}$ denotes connection weights between hidden nodes and output nodes, $b_j$ and $b_k$ denote biases of the hidden node and the output node, respectively, and $q$ is the number of hidden nodes.

TABLE II. RESULTS OF MKPSO, MPANN, DIVACE AND EENCL ON THE AUSTRALIAN CREDIT CARD DATA SET

|  | MKPSO | | MPANN | | DIVACE | | EENCL | |
| --- | --- | --- | --- | --- | --- | --- | --- | --- |
|  | *Training error* | *Testing error* | *Training error* | *Testing error* | *Training error* | *Testing error* | *Training error* | *Testing error* |
| **Mean** | 0.117 | 0.133 | 0.151 | 0.135 | 0.128 | 0.138 | 0.090 | 0.135 |
| **SD** | 0.009 | 0.041 | 0.019 | 0.043 | 0.007 | 0.049 | 0.035 | 0.039 |
| **Min** | 0.105 | 0.073 | 0.123 | 0.072 | 0.116 | 0.073 | 0.076 | 0.087 |
| **Max** | 0.127 | 0.188 | 0.197 | 0.212 | 0.141 | 0.247 | 0.103 | 0.203 |

TALBE III. RESULTS OF MKPSO, MPANN, DIVACE AND EENCL ON THE DIABETES DATA SET

|  | MKPSO | | MPANN | | DIVACE | | EENCL | |
| --- | --- | --- | --- | --- | --- | --- | --- | --- |
|  | *Training error* | *Testing error* | *Training error* | *Testing error* | *Training error* | *Testing error* | *Training error* | *Testing error* |
| **Mean** | 0.217 | 0.219 | 0.231 | 0.223 | 0.220 | 0.227 | 0.205 | 0.234 |
| **SD** | 0.008 | 0.051 | 0.023 | 0.032 | 0.006 | 0.050 | 0.007 | 0.039 |
| **Min** | 0.206 | 0.156 | 0.192 | 0.160 | 0.209 | 0.141 | 0.194 | 0.172 |
| **Max** | 0.228 | 0.299 | 0.263 | 0.277 | 0.232 | 0.313 | 0.217 | 0.297 |

## B. The Encoding Scheme

In our method, we adopt the real coding representation, each particle is encoded as a vector of floating numbers, including all the connected weights and biases of the three-layer NN. The particle can be represented as:

$$pos = (w_{11},...,w_{qn},b_1,...,b_q,w_{11},...,w_{mq},b_1,...,b_m). \quad (14)$$

The dimension of the particle is equal to $n \cdot q + q + q \cdot m + m$.

## C. The Fitness Function

The proposed method aims to minimize the mean squared error (MSE) which is defined as:

$$fitness = \frac{1}{mp}\sum_{k=1}^{m}\sum_{l=1}^{p}(f_k(X^l) - t_k(X^l))^2, \quad (15)$$

where $p$ is the number of input–output pairs of training set, $t$ is the expected output.

## D. The K-means-based Multi-subpopulation PSO

In this work, we regard ensembling NNs as a multimodal optimization problem. Each optimal or near-optimal solution corresponds to a component NN. By employing PSO to deal with a multimodal optimization problem, subpopulations can be formed in the population to explore different regions of the search space, each particle can communicate only with particles in the same subpopulation. Following this paradigm, the use of clustering could help identify and foster these subpopulation, therefore preserving the diversity of component NNs. Here, we propose a *k*-means-based multi-subpopulation PSO, denoted KMPSO, for ensembling NNs. In the proposed KMPSO, we adopt the *k*-means algorithm, which is perhaps the most popularly used clustering algorithm [14,15], to dynamically cluster particles according to their *pbests* at every iteration. In particular, when particles updates its velocities, we replace (10) with

$$v_{id} = wv_{id} + c_1r_1(p_{id} - x_{id}) + c_2r_2(p_{cd} - x_{id}), \; 1 \le c \le k, \quad (16)$$

where $\vec{p}_c$ is the optimal *pbest* in the cluster that current particle belongs to, $k$ is the number of clusters. The procedure of KMPSO is shown in Fig. 1. In the proposed method, the employment of *k*-means clustering algorithm is to keep track of the swarm dynamics: particles in different clusters at early stages of the simulation can end up in the same cluster as they move towards the same local optimum. By contrast, a single cluster can be split into two as some of its particles move to a different optimum. As a result, the difference of component NNs on weight space can be efficiently kept as well as a good trade-off between the diversity and accuracy in the ensemble. After reaching a fixed number of iterations, the best particles from *k* clusters corresponds to *k* component NNs will be used to construct the ensemble.

## V. EXPERIMENTS

In this section, we evaluate the performance of KMPSO on five benchmark classification problems and compare it with Bagging [16], Adaboosting [17], EENCL [18], DIVACE [10] and MPANN[19]. The data sets listed in Table I are from the University of California at Irvine (UCI) machine learning repository [20].

## A. Experimental Settings

In the experiments, we employ *k*-fold cross-validation for deriving a good estimation of generalization error. The input attributes of all data sets are normalized to between 0 and 1 using a linear function. The output attributes of all problems are encoded using a *1-of-m* output representation for *m* classes. The output of NN ensemble is determined as the average output of all component NNs. The proposed KMPSO involves a few control parameters that need to be set. The population size *M* is set as 250 and the maximum number of generations is set as 150. The inertia weights $c_1=c_2=2.0$ and the *w* decreases linearly from 0.9 to 0.2. The connected weights and biases are limited to [-5,5]. The number of hidden neurons for NN is set as 7. The number of clusters $k$ is set as 10.

TABLE IV. RESULTS OF MKPSO, BAGGING AND ADABOOSTING ON FIVE DATA SETS

| Data set\Method | MKPSO | Bagging | Adaboosting |
|---|---|---|---|
| Australian credit card | 0.133 | 0.142 | 0.152 |
| Diabetes | 0.223 | 0.232 | 0.241 |
| Glass | 0.346 | 0.335 | 0.332 |
| Breast cancer | 0.029 | 0.034 | 0.037 |
| heart | 0.158 | 0.184 | 0.211 |

*B. Experimental Results*

Tables II and III present the results of MKPSO, MPANN, DIVACE and EENCL on the Australian credit card and diabetes data sets. We reported the mean, standard deviation, maximum and minimum value of testing and training errors. To be consistent with the literature [10], [18] and [19], the results are obtained by 10-fold cross-validation for the Australian credit card data set and 12-fold cross-validation for the diabetes data set. The error rate in the tables refers to the percentage of wrong classification produced by the trained NN ensemble on the training set and testing set. Table IV shows the results of MKPSO, Bagging and Adaboosting on the five data sets in terms of test error. The results are all obtained by 10-fold cross-validation except the Australian credit card data set which is obtained by 12-fold cross-validation.

It can be observed from the resluts that our proposed method can obtain very competitive results in comparison with related algorithms. For example, on Australian credit card and Diabetes data sets, Table II and III show that KMPSO achieves a better average error rate than MPANN, DIVACE, EENCL, Bagging and Adaboosting. For the other three data sets, Table IV shows KMPSO can also achieve better performance than Bagging and Adaboosting except on the Glass data set. In general, we can conclude that the proposed KMPSO has a good performance in ensembling NNs and outperform related methods.

VI. CONCLUSION

In this work, we treat ensembling NNs as a multimodal optimization problem, and propose a *k*-means-based multi-subpopulation particle swarm optimization for training the NN ensemble. By employing the dynamic cluster training, our method can preserve the diversity of the component NNs on weight space. The experimental results show that the method can deliver satisfactory ensembling NNs, outperforming related methods to be compared.